\documentclass[letterpaper, 10 pt, conference]{IEEEtran}

\usepackage[utf8]{inputenc}
\usepackage[english]{babel}
\usepackage{floatrow}
\usepackage{graphics} 
\usepackage{epsfig} 
\usepackage{mathptmx} 
\usepackage{times} 
\usepackage{cite}
\usepackage{amsmath} 
\usepackage{amssymb}  
\usepackage{graphicx}
\usepackage{mwe}
\usepackage{lipsum}%
\usepackage{booktabs}
\usepackage{multirow}
\usepackage[algoruled,boxed,lined]{algorithm2e}
\usepackage[mathscr]{euscript}
\usepackage{tabularx}
\usepackage{amsfonts}
\usepackage{url}
\usepackage{bm}
\usepackage{adjustbox}
\usepackage{footnote}
\usepackage[para,online,flushleft]{threeparttable}
\usepackage{multirow}
\begin{document}
	
	\title{\LARGE \bf Pedestrian Trajectory Prediction using Context-Augmented Transformer Networks}
		\author{\IEEEauthorblockN{Khaled Saleh}
		\IEEEauthorblockA{Faculty of Engineering and IT\\
			University of Technology Sydney\\
			Sydney, Australia\\
			Email: khaled.aboufarw@uts.edu.au}}
	\maketitle

\begin{abstract}
Forecasting the trajectory of pedestrians in shared urban traffic environments is still considered one of the challenging problems facing the development of autonomous vehicles (AVs). In the literature, this problem is often tackled using recurrent neural networks (RNNs). Despite the powerful capabilities of RNNs in capturing the temporal dependency in the pedestrians' motion trajectories, they were argued to be challenged when dealing with longer sequential data. Thus, in this work, we are introducing a framework based on the transformer networks that were shown recently to be more efficient and outperformed RNNs in many sequential-based tasks. We relied on a fusion of the past positional information, agent interactions information and scene physical semantics information as an input to our framework in order to provide a robust trajectory prediction of pedestrians. We have evaluated our framework on two real-life datasets of pedestrians in shared urban traffic environments and it has outperformed the compared baseline approaches in both short-term and long-term prediction horizons. 		
\end{abstract}

\section{Conclusion }\label{concl}
In this work, we have introduced a novel context-augmented transformer network-based model for the task of pedestrians' trajectory prediction in an urban shared traffic environment. Besides their past trajectories, the proposed model also took into account the contextual information around pedestrians in the scene (i.e. interactions with other agents and the scene's semantics information). The inclusion of such information led the model to achieve robust results on real-life datasets of pedestrians' trajectories in urban shared traffic environments. Moreover, the proposed model has outperformed the compared baseline models from the literature with a large margin over short and long term prediction horizons. 

\bibliographystyle{IEEEtran}
\bibliography{IEEEabrv,Ref}
	
\end{document}